\documentclass[final]{cvpr}

\usepackage{times}
\usepackage{epsfig}
\usepackage{graphicx}
\usepackage{amsmath}
\usepackage{amssymb}
\usepackage{booktabs,tabularx}

\usepackage[pagebackref=true,breaklinks=true,colorlinks,bookmarks=false]{hyperref}
\newcommand{\minisection}[1]{\noindent{\textbf{#1}}}
\DeclareMathOperator*{\argmax}{arg\,max}

\def\cvprPaperID{5538} % *** Enter the CVPR Paper ID here

\def\confYear{CVPR 2021}

\begin{document}

%%%%%%%%% TITLE
\title{A Multiplexed Network for End-to-End, Multilingual OCR}

\author{
Jing Huang\quad Guan Pang\quad Rama Kovvuri\quad Mandy Toh\quad Kevin J Liang\\ Praveen Krishnan\quad Xi Yin\quad Tal Hassner
\\
Facebook AI
\\
{\tt\small 
\{jinghuang,gpang,ramakovvuri,mandytoh,kevinjliang,pkrishnan,yinxi,thassner\}@fb.com}
}

\maketitle

\begin{abstract}
   Recent advances in OCR have shown that an end-to-end (E2E) training pipeline that includes both detection and recognition leads to the best results.
   However, many existing methods focus primarily on Latin-alphabet languages, often even only case-insensitive English characters. 
   In this paper, we propose an E2E approach, Multiplexed Multilingual Mask TextSpotter, that performs script identification at the word level and handles different scripts with different recognition heads, all while maintaining a unified loss that simultaneously optimizes script identification and multiple recognition heads. Experiments show that our method outperforms the single-head model with similar number of parameters in end-to-end recognition tasks, and achieves state-of-the-art results on MLT17 and MLT19 joint text detection and script identification benchmarks. We believe that our work is a step towards the end-to-end trainable and scalable multilingual multi-purpose OCR system. Our code and model will be released.
\end{abstract}

\section{Introduction}
Reading text in visual content has long been a topic of interest in computer vision, with numerous practical applications such as search, scene understanding, translation, navigation, and assistance for the visually impaired.
In recent years, advances in deep learning have led to dramatic improvements of Optical Character Recognition (OCR), allowing reading text in increasingly diverse and challenging scene environments with higher accuracy than ever before.
A common approach is to decompose the task into two sub-problems: text detection, the localization of text in visual media, and text recognition, the transcription of the detected text.
While these two components were traditionally learned separately, recent works have shown that they can be learned jointly, with benefits to both modules. 

As the most commonly spoken language in the world~\cite{english2019} and a \textit{lingua franca} for research, the English language has been the focus of many public OCR benchmarks~\cite{lucas2003icdar, wang2010word, karatzas2013icdar, karatzas2015icdar, veit2016coco, textocr} and methods~\cite{liao2020mask, liu2020abcnet, qiao2020text, qin2019towards}.
However, English (and other Latin alphabet languages) represent only a fraction of the languages spoken (and written) around the world. OCR technology is also used to study forgotten languages and ancient manuscripts, where alphabets and script styles can vary enormously~\cite{hassner2012computation,hassner2014digital}. 
Thus, developing OCR capabilities in other languages is also important to ensure such technologies are accessible to everyone.
Additionally, because of the increasing interconnectedness of the world and its cultures, it is important to develop OCR systems capable of recognizing text from multiple languages co-occurring in the same scene.

While many concepts and strategies from OCR on English text can be adapted to other languages, developing multilingual OCR systems is not completely straightforward.
Naively training a separate system for each language is computationally expensive during inference and does not properly account for predictions made for other languages.
Furthermore, previous works~\cite{qin2019towards, liao2020mask} have shown that jointly learning text detection and text recognition modules is mutually beneficial; separate models lose out on the potential benefits of a shared text detection module.
On the other hand, learning a unified model with single recognition head also presents problems.
While uncased English only has 26 characters, many Asian languages like Chinese, Japanese, and Korean have tens of thousands of characters. Different languages/scripts can also have very different word structures or orientations. For example, vertically written text is far more common in East Asian languages like Chinese, Japanese and Korean than in Western languages, and  characters in Arabic and Hindi are usually connected to each other.
This variability in the number of characters as well as the wide variability in script appearance characteristics mean it is highly unlikely that a single architecture can capably maximize accuracy and efficiency over all languages/scripts, and any imbalances in the training data may result in significantly different performances between languages.

Given these challenges, we present a blend of these two approaches, incorporating each one's advantages while mitigating their faults. 
Specifically, we propose a single text detection module followed by a text recognition head for each language, with a multiplexer routing the detected text to the appropriate head, as determined by the output of a Language Prediction Network (LPN).
This strategy can be seen as analogous to human perception of text.
Locating the words of most languages is easy even without knowing the language, but recognizing the actual characters and words requires special knowledge: language/script identification typically precedes recognition.

Notably, this multiplexer design has important implications for real-world text spotting systems.
Having language-specific text recognition heads allows custom design of the architecture depending on the difficulty and characteristics of each language, while still sharing and jointly learning the same text detection trunk.
New languages can also be easily added to the system without re-training the whole model and worrying about affecting the existing languages.

Our contributions can be summarized as follows:

\begin{itemize}
\item We propose an end-to-end trainable multiplexed OCR model that can automatically pick the best recognition head for the detected words.
\item We propose a language prediction network using masked pooled features as input and an integrated loss function with the recognition heads.
\item We design a training strategy that takes advantage of the proposed losses, allows for easy extension to new languages and addresses the data imbalance problem.
\item We empirically show that the multiplexed model consistently outperforms single-head model and is less prone to training data distribution bias.
\end{itemize}

\section{Related work}
Text spotting is commonly broken down into two sub-tasks: text detection and text recognition.
In scenes with multiple languages, script identification is also necessary, either explicitly by learning a classification model or implicitly as a byproduct of text recognition.
While these three sub-tasks were often considered individually and then chained together in the past, end-to-end methods seeking to learn all at once have recently become popular.
We give a brief overview of relevant works below; see \cite{long2020scene} for a more thorough treatment.

\subsection{Text detection}
Text detection is commonly the first stage of understanding text content in images.
Early approaches typically consisted of human-engineered features or heuristics, such as connected components~\cite{jain1998automatic, neumann2012real, yin2013robust} or sliding windows~\cite{lee2011adaboost}.
The promise of early deep learning models~\cite{krizhevsky2012imagenet} led to some of these strategies being combined with convolutional networks~\cite{wang2012end, huang2014robust}, and as convolutional networks proved successful for object detection~\cite{girshick2014rich, ren2015faster, he2017mask}, more recent approaches have almost exclusively been using deep detection models~\cite{tian2016detecting}. 
Given the various orientations and shapes that text can take, further refinements have focused on making text detection rotation invariant~\cite{jiang2017r2cnn} or switched from rectangular bounding boxes to more flexible segmentation masks~\cite{liao2019mask, qin2019towards}.
Character-level detection with weakly supervised learning of word-level annotations has also been shown effective~\cite{baek2019character}.

\subsection{Text recognition}
Once text has been localized through detection, the region is often cropped and then fed to a text recognition system to be read as a character/word sequence.

Like text detection, text recognition methods have a long history predating the popular use of deep learning~\cite{chen2004automatic, nomura2005novel, shi2013scene, rodriguez2013label}, but most recent methods use neural networks.
Connectionist temporal classification (CTC)~\cite{graves2006connectionist} methods use recurrent neural networks to decode features (recently mostly convolutional) into an output sequence~\cite{su2014accurate, he2016reading}.
Another common framework for text recognition is the Seq2Seq encoder-decoder framework~\cite{sutskever2014sequence} that is often combined with attention~\cite{bahdanau2014neural}, which is used by~\cite{lee2016recursive, shi2016robust}.
\cite{jaderberg2016reading} frames the problem as a $V$-way image classification problem, where $V$ is the size of a pre-defined vocabulary.

\subsection{Script identification}
Text spotting in multilingual settings often requires script identification to determine a language for text recognition.
Early works focused on identifying the language of scripts in simple environments like documents~\cite{hochberg1997automatic, tan1998rotation, busch2005texture}, primarily with traditional pre-deep learning methods.

As with other vision tasks, convolutional architectures proved especially effective~\cite{sharma2015icdar2015, gomez2017improving}.
Script identification in natural scenes began with Shi \etal~\cite{shi2015automatic}, who cropped and labeled text in images from Google Street View, then trained convolutional neural networks with a specialized multi-stage pooling layer for classification; the authors achieved further gains in accuracy with densely extracted local descriptors combined with discriminitive clustering.
Fujii \etal~\cite{fujii2017sequence} proposed a line-level script identification method casting the problem as a sequence-to-label problem. 
E2E-MLT~\cite{buvsta2018e2e} is a multilingual end-to-end text spotting system that forgoes script identification and performs text recognition directly. They proposed to use a CNN to classify the script at the cropped-word level that preserves the aspect ratio.
Another common approach for script identification comes after the text recognition step, which infers the language by identifying the most frequent language occurrences of the characters in the text~\cite{baek2020character}. The resulting more challenging text recognition task leads to somewhat hampered model performance.
We find that performing script identification in our proposed Language Prediction Network (LPN), with the masked pooled features of the detected words as input, to multiplex the text recognition heads leads to significantly higher accuracy for the script identification task, compared to the majority-vote approach.

\begin{figure*}[ht]
    \centering
    \includegraphics[width=0.98\linewidth]{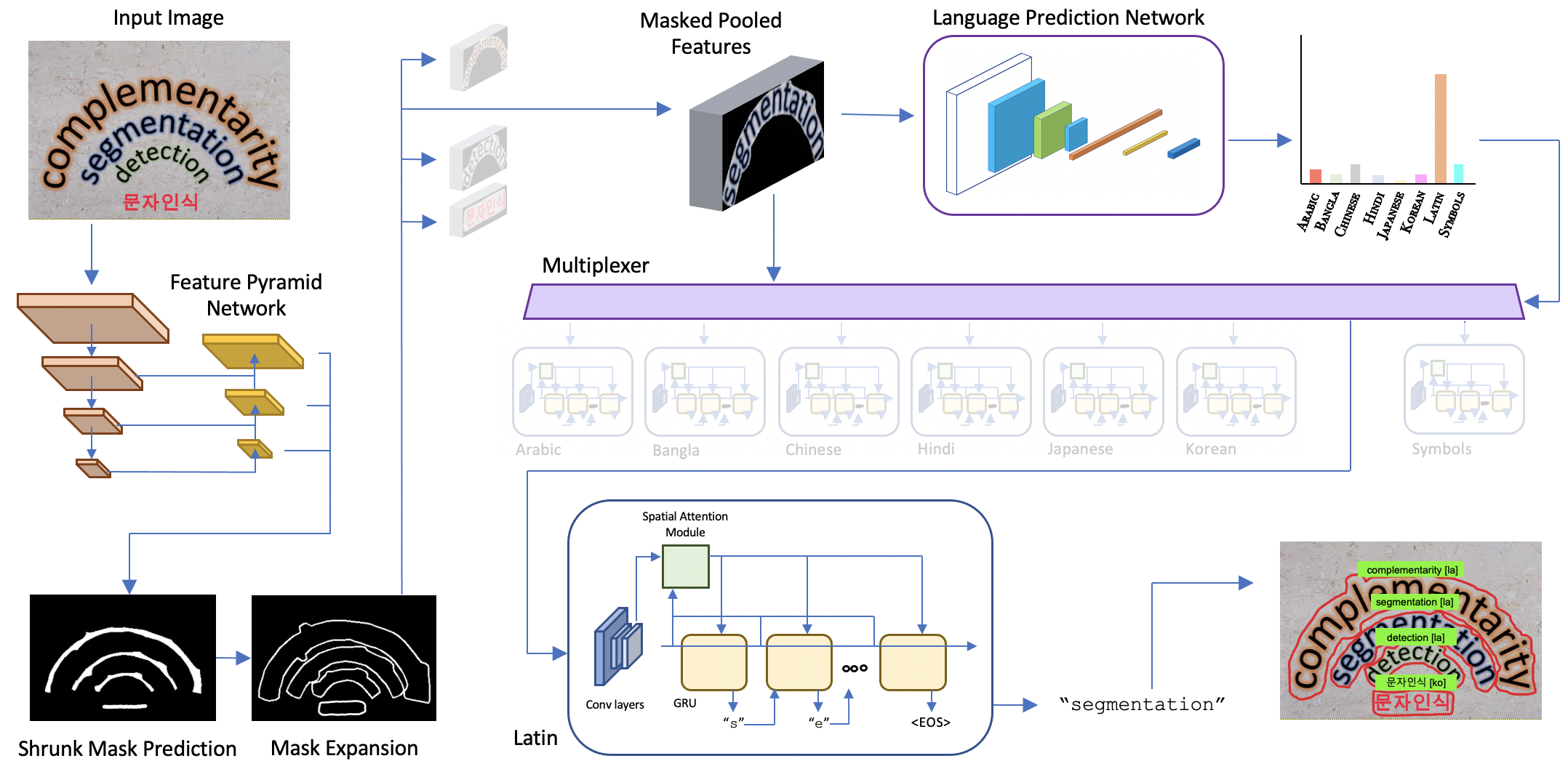}
    % \vspace{-2mm}
    \caption{\textbf{M3 TextSpotter.} The proposed M3 TextSpotter shares the same detection and segmentation trunk with Mask TextSpotter v3~\cite{liao2020mask}, but incorporates a novel Language Prediction Network (LPN).
    The output of the LPN then determines which script's recognition head the multiplexer selects.}
    \label{fig:detection_segmentation_trunk}
    % \vspace{-6mm}
\end{figure*}

\subsection{Text spotting}
While many early works focused on one of the aforementioned tasks in isolation, end-to-end (E2E) text spotting systems have also been proposed.

Some learn text detection and recognition submodules separately, linking the two independent systems together for the final product~\cite{neumann2012real, jaderberg2016reading, liao2016textboxes}. 
However, the learning tasks of text detection and recognition are mutually beneficial: recognition can provide additional feedback to detection and remove false positives, while detection can provide augmentations for recognition. 
As such, recent works learn these two jointly~\cite{liu2018fots, qin2019towards, liao2019mask, liao2020mask}. 
E2E-MLT~\cite{buvsta2018e2e} proposed an E2E text spotting method evaluated on multilingual settings, but does not explicitly incorporate any specific model components adapted for multiple languages, instead dealing with characters from all languages in the same recognition head; this is the approach taken by most E2E systems for multilingual settings like ICDAR-MLT~\cite{nayef2017icdar2017, nayef2019icdar2019} and CRAFTS~\cite{baek2020character}.

\section{Methodology}
The multiplexed model shares the same detection and segmentation modules as Mask TextSpotter V3~\cite{liao2020mask} (Figure \ref{fig:detection_segmentation_trunk}). A ResNet-50~\cite{he2016deep} backbone with a U-Net structure~\cite{ronneberger2015u} is used to build the Segmentation Proposal Network (SPN). Similar to~\cite{liao2020mask, wang2019shape}, the Vatti clipping algorithm~\cite{vati1992generic} is used to shrink the text regions with a shrink ratio $r$ to separate neighboring text regions. Once the segmentation proposals are generated, hard RoI masking~\cite{liao2020mask} is used to suppress the background and neighboring text instances.

The recognition model for Mask TextSpotter V3~\cite{liao2020mask} comprises of a Character Segmentation Module and a Spatial Attention Module~\cite{liao2019mask} adapted for text recognition. We only use the Spatial Attention Module in our model due to the following reasons: (1) using both modules does not scale when expanding the character set from 36 to 10k; (2) the Character Segmentation Module requires character-level annotations to supervise the training and the order of the characters cannot be obtained from the segmentation maps; (3) in our experiments on Latin-only model, disabling the Character Segmentation Module has a minimal effect on the final recognition results.

To extend the model from Latin-only to multilingual, there are two directions: (1) treat all languages and characters as if they belong to the same language with all characters, and use a single recognition head to handle all of them; (2) build separate recognition heads to handle words from different languages, and then pick/combine the predictions from them. We choose approach (2) since it is much more flexible when we train the model without worrying about data imbalance across different languages, and has greater potential for future extension, e.g., incorporating language model into the recognition.

\subsection{Language prediction network}
\begin{figure}[b]
    \centering
    \includegraphics[width=0.98\linewidth]{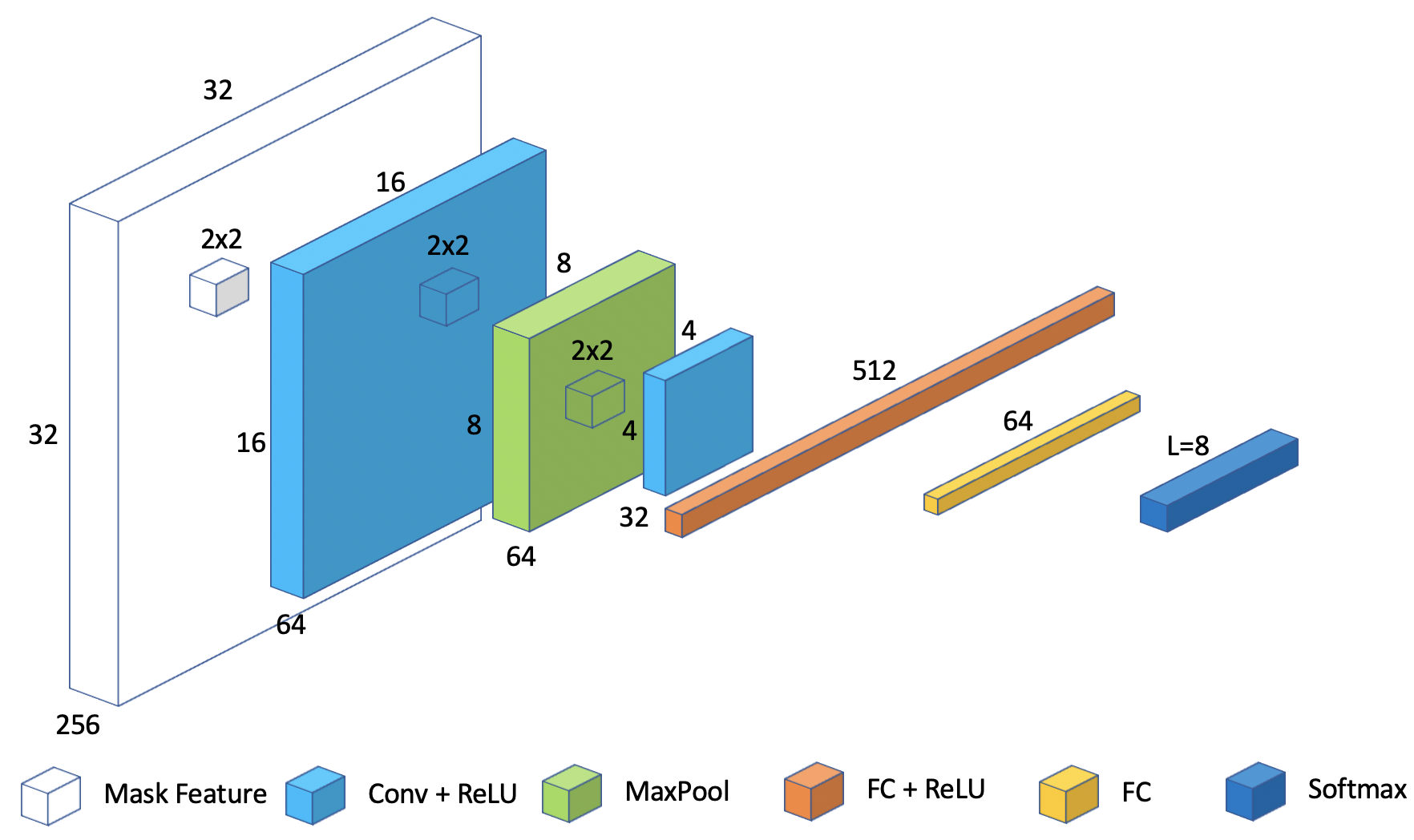}
    % \vspace{-2mm}
    \caption{\textbf{Language Prediction Network.} In this figure $L = N_{lang} = 8$, denoting the eight scripts (Arabic, Bengali, Chinese, Hindi, Japanese, Korean, Latin and Symbol) supported by our default model in this paper.}
    \label{fig:language_prediction_network}
    % \vspace{-6mm}
\end{figure}
To automatically select the recognition module appropriate for a given script, we propose a Language Prediction Network (LPN), as detailed in Figure \ref{fig:language_prediction_network}. The input of the LPN is the masked pooled feature from the detection and segmentation modules, with size $256 \times 32 \times 32$. We apply a standard classification network with two $2 \times 2$ convolutional layers with rectified linear unit (ReLU) activations and a $2 \times 2$ max pooling layer in between, followed by two fully connected (FC) layers with a ReLU in between. This network produces an output vector of size $L$, which can be converted into probabilities using a Softmax, where $L = N_{lang}$ is the number of language classes we would like the model to support.

Note that in practice, the number of different recognition heads in the model $N_{rec}$ does not necessarily have to equal the number of supported languages $N_{lang}$, particularly in the case of shared alphabets. 
For example, the Latin alphabet is used for Germanic (e.g. English, German) and Romance (e.g. French, Italian) languages, meaning LPN predictions for any of these languages can be routed to a singular Latin recognition head. 
For simplicity, in the following sections we assume $N_{lang} = N_{rec}$.

Finally, we note that previous work suggested network decision making mechanisms~\cite{wu2017facial}. These methods were proposed for very different applications than the one considered here. Importantly, the decision mechanisms they described were not network-based and so not end-to-end trainable with other network components. 

\subsection{Multiplexer with disentangled loss}

Since a few datasets (e.g., MLT~\cite{nayef2017icdar2017}) provide ground truth annotations for the language of a particular text, we can train both the LPN and the recognition heads in parallel with a disentangled loss, i.e., computing the loss terms for each of the heads and the LPN in parallel and then directly adding them up:
\begin{equation} \label{eq:disentagled}
    L_{disentangled} = \alpha_{lang} L_{lang} + \sum_{r \in R} \alpha_{seq(r)} L_{seq(r)}
\end{equation}
where $L_{lang}$ is the loss for LPN, $R$ is the set of recognition heads, and $L_{seq(r)}$ is the loss for the recognition head $r$. $\alpha_{lang}$ and $\alpha_{seq(r)}$ are weighting hyper-parameters. In our experiment, we set $\alpha_{lang}=0.02$ in the first few thousands of iterations in the first training stage (\ref{sec:training_strategy}) and $\alpha_{lang}=1$ after that; we use $\alpha_{seq(r)}=0.5$ for all recognition heads throughout the training.

Language prediction is a standard $N$-way classification problem, so the language prediction loss in Equation~\ref{eq:disentagled} can be computed using a cross entropy loss:
\begin{equation}
    L_{lang} = - \sum_{l=1}^{N_{lang}} I(l=l_{gt}) \log p(l),
\end{equation}
where $I(l=l_{gt})$ is the binary indicator (0 or 1) if the language matches the ground truth, and $p(l)$ is the probability inferred by the LPN that the word belongs to language $l$.

Similar to \cite{liao2019mask}, we use the negative log likelihood as the text recognition loss $L_{seq(r)}$:
\begin{equation}
    L_{seq} = - \frac{1}{T} \sum_{t=1}^{T} \log p(y_t=c_t),
\end{equation}
where $p(y_t=c_t)$ is the predicted probability of character at position $t$ of the sequence, and $T$ is the length of the sequence of character labels. We use $T=32$ for all the recognition heads in this paper, but it can be customized to account for different distributions of word length across the languages - for example, since there's typically no space between the Chinese and Japanese words, we can use bigger $T$ for these languages.

To compute $L_{seq}(r)$, i.e., $L_{seq}$ for different recognition heads that support different character sets, we need to ignore the unsupported characters in the loss computation:

\begin{equation}
    L_{seq(r)} = - \frac{1}{T} \sum_{t=1}^{T} I(c_t \in C_r) \log p(y_t=c_t),
    \label{eqn:seq_loss_ignoring_unsupported_chars}
\end{equation}
where $C_r$ is the character set supported by recognition head $r$, $c_t$ is the ground truth character at step $t$,  $I(c_t \in C_r) = 1$ if $c_t$ is supported and $I(c_t \in C_r) = 0$ if $c_t$ is not supported.

\subsection{Multiplexer with integrated loss}

While the multiplexer with disentangled loss could serve as a good initialization for model training, such an approach has a few limitations. First, the training of the language predictor requires explicit ground truth annotations of the language at the word level, which can be inaccurate and is not always available outside of curated datasets. Secondly, the disentangled total loss does not reflect the actual prediction of the model at inference time, especially when there are shared characters across multiple recognition heads. Finally, despite having a mechanism to ignore labels, it is counter-productive to train the recognition heads for the wrong language with unsupported words.

To address these problems, we propose an integrated loss that combines results from the language prediction head and the recognition heads during training. To enforce consistency between training and testing, we can use a hard integrated loss:

\begin{equation}
    L_{hard-integrated} = \alpha_{seq(r)} L_{seq(\argmax_{1 \leq l \leq N_{rec}} p(l))}
\end{equation}

\begin{figure*}[t]
    \centering
    \includegraphics[width=0.98\linewidth]{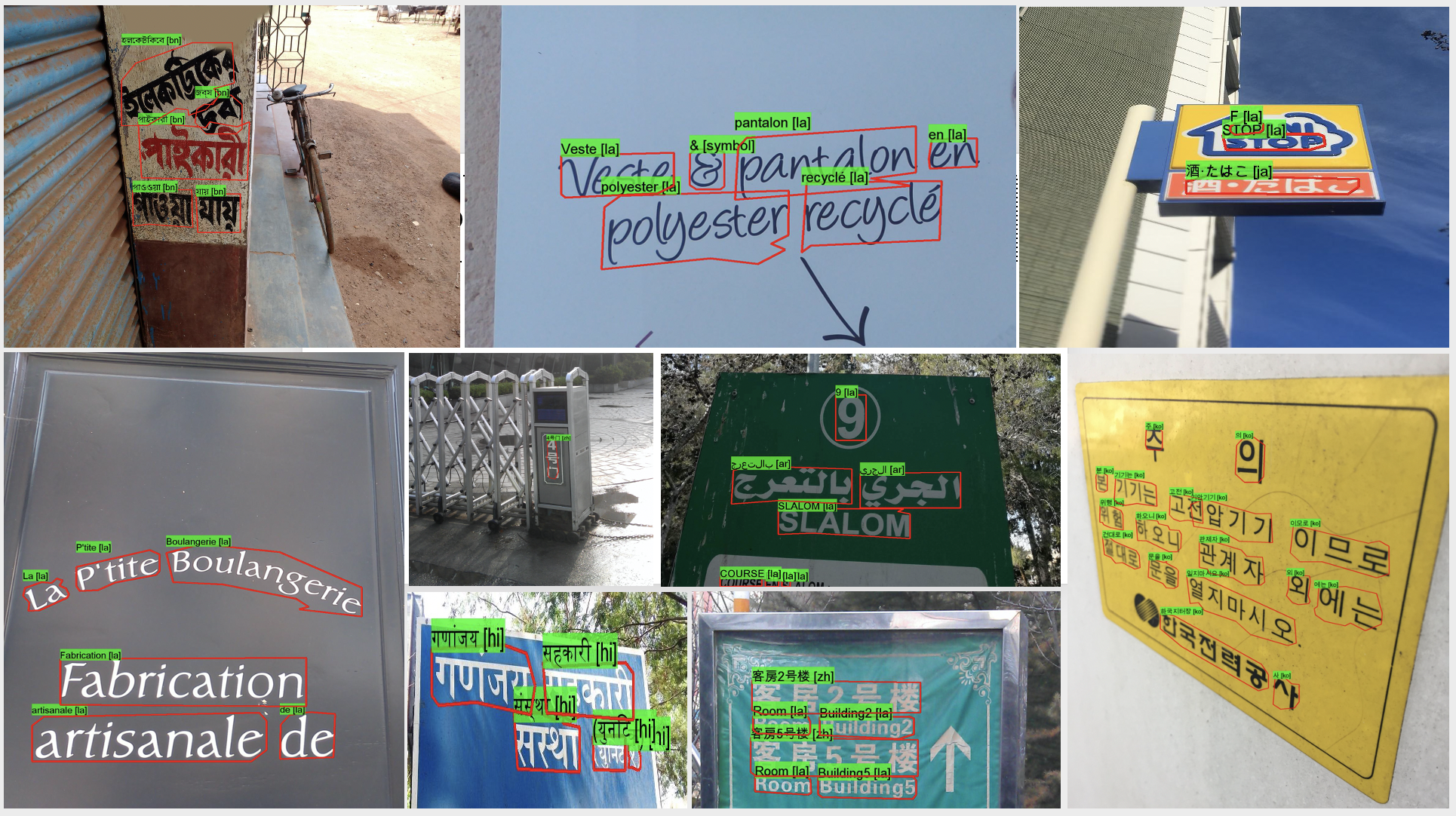}
    % \vspace{-2mm}
    \caption{\textbf{Qualitative results on MLT19.} The polygon masks predicted by our model are shown over the detected words. The transcriptions from the selected recognition head and the predicted languages are also rendered in the same color as the mask. The language code mappings are: ar - Arabic, bn - Bengali, hi - Hindi, ja - Japanese, ko - Korean, la - Latin, zh - Chinese, symbol - Symbol.}
    \label{fig:vis2_mlt19test}
    % \vspace{-6mm}
\end{figure*}

With a hard integrated loss, we pick exactly one recognition head for each word, selecting and using the loss of the head that has the maximum probability as predicted by the language prediction network. This loss better matches the operation of the text spotting system during inference and avoids involving irrelevant recognition heads during training. Our ablation study (Section~\ref{sec:integrated_loss_ablation}) shows that it outperforms an alternative soft integrated loss (Equation~\ref{eq:soft_integarted_loss}).

Note that directly using the default sequence recognition loss (Equation \ref{eqn:seq_loss_ignoring_unsupported_chars}) in the integrated losses does not work due to the handling of the unsupported characters: unsupported characters will always contribute $0$ to the loss while supported characters contribute a positive value to the total loss, no matter how good the actual prediction is. To resolve this problem, we can assign a large penalty factor $\beta$ to unsupported characters:
\begin{equation}
    L_{seq(r)} = - \frac{1}{T} \sum_{t=1}^{T} [I(c_t \in C_r) \cdot \log p(y_t=c_t) + I(c_t \notin C_r) \cdot \beta]
    \label{eqn:seq_loss_penalizing_unsupported_chars}
\end{equation}
We set the penalty to $\beta=-12$ in our experiments.

\section{Experiments}
We validate the effectiveness of our multilingual multiplexer design with a series of experiments, evaluating the proposed Multiplexed Mask TextSpotter on multilingual scene text from the MLT17~\cite{nayef2017icdar2017} and MLT19~\cite{nayef2019icdar2019} datasets.
In addition to these two datasets, we also take advantage of several other public OCR datasets for training.
We report results for text detection, end-to-end script identification, and end-to-end multilingual recognition tasks.
We also show the results of an ablation study comparing our multiplexed multi-headed approach with a single combined recognition head approach.

\subsection{Datasets}

\minisection{ICDAR 2017 MLT dataset (MLT17)}~\cite{nayef2017icdar2017} was introduced as a part of ICDAR 2017 Robust Reading Competition for the problem of multi-lingual text detection and script identification. It contains 7200 training, 1800 validation and 9000 test images in 9 languages representing 6 different scripts equally. The dataset contains multi-oriented scene text that is annotated using quadrangle bounding boxes.

\minisection{ICDAR 2019 MLT dataset (MLT19)}~\cite{nayef2019icdar2019} was introduced as a part of ICDAR 2019 Robust Reading Competition extending ICDAR 2017 MLT dataset for the problem of multi-lingual text detection and script identification. It contains 10000 training and 10000 test images in 10 languages representing 7 different scripts. The dataset also contains multi-oriented scene text that is annotated using quadrangle bounding boxes. It also provides a synthetic dataset (SynthTextMLT)~\cite{buvsta2018e2e} that provides $\sim$273k synthetic data in 7 scripts. There are many errors for Hindi images in SynthTextMLT, so we filtered out any Hindi images containing non-Hindi characters (likely errors) when using it.

\minisection{Total-Text dataset}~\cite{ch2017total}, presented at ICDAR 2017 is a comprehensive scene text dataset for text detection and recognition. It contains 1255 training and 300 test images in English language. The dataset contains wide variety of horizontal, multi-oriented and curved text annotated at word-level using polygon bounding boxes.

\minisection{ICDAR 2019 ArT dataset (ArT19)}~\cite{chng2019icdar2019} was introduced as a part of ICDAR 2019 Robust Reading Competition. It contains 5603 training and 4563 test images in English and Chinese languages. The dataset is a combination of Total-Text~\cite{ch2017total} and SCUT-CTW1500~\cite{yuliang2017detecting} datasets. The dataset contains highly challenging arbitrarily shaped text that is annotated using arbitrary number of polygon vertices. Since this dataset contains the testing images from Total Text, we deliberately filtered them out in our training so that our model weights remain valid for future training/evaluation on the Total Text benchmark.

\minisection{ICDAR 2017 RCTW dataset (RCTW17)}~\cite{shi2017rctw} was introduced as a part of ICDAR 2017 Robust Reading Competition on Reading Chinese Text in the Wild. It contains 8034 train and 4229 test images, focusing primarily on scene text in Chinese.

\minisection{ICDAR 2019 LSVT dataset (LSVT19)}~\cite{sun2019lsvt} was introduced as a part of ICDAR 2019 Robust Reading Competition on Large-scale Street View Text with Partial Labeling. It is one of the largest OCR datasets, containing 30000 train and 20000 test images. The dataset is primarily street view text in Chinese, but also has about 20\% of its labels in English words.

\minisection{ICDAR 2013 dataset (IC13)}~\cite{karatzas2013icdar} was introduced as part of the ICDAR 2013 Robust Reading Competition. It contains 229 training and 233 test images in English language. The dataset contains high-resolution, horizontal text annotated at word-level using rectangular bounding boxes.

\minisection{ICDAR 2015 dataset (IC15)}~\cite{karatzas2015icdar} was introduced as part of the ICDAR 2015 Robust Reading Competition. It contains 1000 training and 500 test images in English language. The dataset contains multi-oriented scene text annotated at word-level using quadrangle bounding boxes.

\begin{table}[t]
    \scriptsize
    \centering
    \caption{\textbf{Parameter number comparison between multiplexed model vs. single-head model.} The total number of the multiplexed model is the sum of the parameter numbers for each individual recognition heads as well as the LPN. The parameters for detection, segmentation and mask feature extraction are not included here.}
    \vspace{2mm}
    \begin{tabularx}{0.98\linewidth}{ccccc}
    \toprule
    Head & Charset Size & Embed Size & Hidden Size & Parameters \\
    \hline
    Arabic & 80 & 100 & 224 & 1.15M \\
    Bengali & 110 & 100 & 224 & 1.16M \\
    Chinese & 5200 & 200 & 224 & 3.36M \\
    Hindi & 110 & 100 & 224 & 1.16M \\
    Japanese & 2300 & 200 & 224 & 2.13M \\
    Korean & 1500 & 200 & 224 & 1.79M \\
    Latin & 250 & 150 & 256 & 1.49M \\
    Symbol & 60 & 30 & 64 & 0.21M \\
    \hline
    LPN & - & - & - & 0.11M \\
    \hline
    Multiplexed & - & - & - & 12.5M \\
    \hline
    Single-Head & 9000 & 400 & 512 & 12.6M \\
    \bottomrule
    \end{tabularx}
    \label{tab:parameter_number_comparison}
    \vspace{-2mm}
\end{table}

\subsection{Training strategy}
\label{sec:training_strategy}

Prior to training, we go through the annotations of the aforementioned datasets to obtain a character set (charset) for each of the eight scripts. Since digits and common punctuation marks appear in all languages, we append them to all character sets. The final number of characters for each recognition head are listed in the column Charset Size of Table \ref{tab:parameter_number_comparison}. The choice of the parameters is based on the following heuristics: we use bigger embedding size (200) for Chinese/Japanese/Korean recognition heads, as they have much bigger character sets; Latin has relatively larger character sets than the remaining scripts as well as much more data, so we use an embedding size of 150 and a bigger hidden layer size of 256 to capture more sequential relationship among the characters. We order each character set by the frequencies of individual characters and map them to consecutive indices for each recognition head, respectively.

We initialize the detection, segmentation, and mask feature extraction weights from the officially published weights released by Mask TextSpotter v3~\cite{liao2020mask}. For the recognition weights, we discard the Character Segmentation Module, and initialize each of the individual heads with the sequence recognition head with spatial attention module with zero-padding or narrowing, since the dimensions of character sizes, embed layer sizes, and hidden layer sizes are different from the original weights.

\begin{table}[t]
    \scriptsize
    \centering
    \caption{\textbf{Quantitative detection results on MLT17.} Note that (1) our model supports Hindi, which is not required by MLT17. (2) CharNet H-88 has 89.21M parameters, which is 3x heavier than CharNet R-50 that is more comparable to our backbone.}
    % \vspace{-2mm}
    \begin{tabularx}{0.9\linewidth}{lX*{2}X}
    \toprule
    Method & F & P & R \\ 
    \midrule
    Lyu et al.~\cite{lyu2018multi} & 66.8 & 83.8 & 55.6 \\
    FOTS~\cite{liu2018fots} & 67.3 & 81 & 57.5 \\
    CRAFT~\cite{baek2019character} & 73.9 & 80.6 & 68.2 \\
    CharNet R-50~\cite{xing2019charnet} & 73.42 & 77.07 & 70.10 \\
    CharNet H-88~\cite{xing2019charnet} & \textbf{75.77} & 81.27 & \textbf{70.97} \\
    \hline
    Multiplexed TextSpotter & 72.42 & \textbf{85.37} & 62.88 \\ 
    \bottomrule
    \end{tabularx}
    \label{tab:mlt17_task1}
    %\vspace{-4mm}
\end{table}

In the first stage of training, we train the model end-to-end using the disentangled loss on datasets (MLT and SynthTextMLT) with ground truth annotations for languages. 
This leads to quicker initial convergence of both the LPN and the recognition heads, as a randomly initialized LPN is unlikely to be able to correctly identify scripts, severely hampering each of the recognition heads from learning, and poorly performing recognition heads deprives the LPN of feedback on its routing.

In the second stage of training, we switch to the hard integrated loss. 
This enables training on all datasets, as explicit ground truth language annotations are no longer necessary for learning.

\begin{table*}[ht]
    \scriptsize
    \centering
    \caption{\textbf{Quantitative detection results on MLT19 with language-wise performance.} All numbers are from the official ICDAR19-MLT website except CRAFTS (paper), which comes from their paper~\cite{baek2020character}. }
    % \vspace{-2mm}
    \begin{tabularx}{0.9\linewidth}{c|cccc|cccccccc}
    \toprule
    Method & F & P & R & AP & Arabic & Latin & Chinese & Japanese & Korean & Bangla & Hindi \\ 
    \midrule
    PSENet~\cite{wang2019shape} & 65.83 & 73.52 & 59.59 & 52.73 & 43.96 & 65.77 & 38.47 & 34.47 & 51.73 & 34.04 & 47.19\\
    RRPN~\cite{ma2018arbitrary} & 69.56 & 77.71 & 62.95 & 58.07 & 35.88 & 68.01 & 33.31 & 36.11 & 45.06 & 28.78 & 40.00\\
    CRAFTS~\cite{baek2020character} & 70.86 & 81.42 & 62.73 & 56.63 & 43.97 & 72.49 & 37.20 & 42.10 & 54.05 & 38.50 & \textbf{53.50}\\ 
    CRAFTS (paper)~\cite{baek2020character} & \textbf{75.5} & 81.7 & \textbf{70.1} & - & - & - & - & - & - & - & -\\
    \hline
    Single-head TextSpotter~\cite{liao2020mask}  & 71.10 & 83.75 & 61.76 & 58.76 & 51.12 & \textbf{73.56} & 40.41 & 41.22 & 56.54 & 39.68 & 49.00 \\ 
    Multiplexed TextSpotter & 72.66 & \textbf{85.53} & 63.16 & \textbf{60.46} & \textbf{51.75} & 73.55 & \textbf{43.86} & \textbf{42.43} & \textbf{57.15} & \textbf{40.27} & 51.95\\ 
    \bottomrule
    \end{tabularx}
    \label{tab:mlt19_task1}
    \vspace{-3mm}
\end{table*}

For the third and final stage of training, we freeze most of the network, including detection, segmentation, mask feature extraction, language prediction networks and all but one individual recognition heads, and train the specific recognition head with only data from this one script. This step would have been impossible if we use the single combined head for all languages, and it greatly resolves the data imbalance problem across different languages.

\subsection{Ablation study} \label{sec:ablation}

\minisection{Multiplexed model vs. single combined head.} In order to make a fair comparison between the multiplexed model versus a single-head model, we estimated the individual as well as the total number of characters to be supported by the eight types of scripts, and adjusted the embedding and hidden layer sizes such that the total number of parameters are roughly the same between the multiplexed model (including the Language Prediction Network) and the single combined-head model (Table \ref{tab:parameter_number_comparison}).

Note that for the multiplexed model, we use a limited set of hidden layer sizes and embedding layer sizes in our experiment. However, these hyper-parameters, or even the underlying architectures, can be further customized based on the importance, difficulty and characteristics of the scripts/languages.

From the experiments in detection (Table \ref{tab:mlt17_task1}, Table \ref{tab:mlt19_task1}), and end-to-end recognition (Table \ref{tab:mlt19_task4}), we can see that the multiplexed model consistently outperforms the single combined head model. Moreover, the multiplexed model can provide extra signal of language identification results (Table \ref{tab:mlt17_task3} and Table \ref{tab:mlt19_task3}) based on visual information, which is not directly available from the single-head model. There are some approaches that can infer the language during post-processing, however, they will need extra language model information to identify the language if the characters are not exclusive to certain languages.

\vspace{2mm}
\minisection{Hard vs. soft integrated loss.} \label{sec:integrated_loss_ablation} There is a legitimate concern on whether the hard integrated loss is differentiable. Instead of using the $\argmax$, we can also employ a soft relaxation to directly multiply the probabilities of each language with the loss from each recognition head and sum them up, yielding the following soft integrated loss function:

\begin{equation}%\vspace{-4mm}
    L_{soft-integrated} = \sum_{r=1}^{N_{rec}} p(r) \cdot \alpha_{seq(r)} L_{seq(r)}
    \label{eq:soft_integarted_loss}
\end{equation}

The hard integrated loss can be seen as a special case of soft integrated loss, where only one of $p(r)$ is $1$ while all others are $0$. In our experiments, however, using hard integrated loss gives about {10\%} better results in terms of H-mean than using soft integrated loss under the same number of iterations. This can be explained by that the hard integrated loss aligns more with the expected behavior of the model during inference time.

\subsection{Text detection task}

Text detection precision (P) and recall (R) for our Multiplexed TextSpotter and several baselines for on MLT17~\cite{nayef2017icdar2017} and MLT19~\cite{nayef2019icdar2019} are shown in Tables~\ref{tab:mlt17_task1} and \ref{tab:mlt19_task1}, respectively.
Note that our model is not fine-tuned on MLT17, which contains one fewer language (Hindi), but still manages to achieve comparable results as other SOTA methods and the highest precision.

In Table \ref{tab:mlt19_task1}, we also show the language-wise F-measure results. Our method beats entries from all published methods on the leaderboard including CRAFTS~\cite{baek2020character}, except the result reported in the paper version of CRAFTS~\cite{baek2020character} is higher. For language-wise evaluation, our method shows the best result for all languages except Hindi, with especially large improvements in Arabic, Chinese, and Korean. Interestingly, we find that the single-head Mask TextSpotter performs slightly better than the Multiplexed Mask TextSpotter in Latin. 
We hypothesize that this is because of the higher prevalence of Latin words in the MLT dataset, due to the inclusion of 4 languages with Latin alphabets: English, French, German, and Italian. Thus, the single-head model greatly favoring Latin leads to stronger Latin performance, to the detriment of the other languages.
This demonstrates that the single-head model is more vulnerable to training data distribution bias.
By contrast, the Multiplexed Mask TextSpotter achieves more equitable performance due to its design.

\subsection{End-to-end script identification task}
\begin{table}[t]
    \scriptsize
    \centering
    \caption{\textbf{Joint text detection and script identification results on MLT17.} Note that our general model supports Hindi, which is not required by MLT17, but still achieves the best result.}
    % \vspace{-2mm}
    \begin{tabularx}{0.9\linewidth}{lX*{3}X}
    \toprule
    Method & F & P & R & AP \\ 
    \midrule
    E2E-MLT~\cite{buvsta2018e2e} & 58.69 & 64.61 & 53.77 & - \\
    CRAFTS~\cite{baek2020character} & 68.31 & 74.52 & \textbf{63.06} & 54.56 \\ 
    \hline
    Multiplexed TextSpotter & \textbf{69.41} & \textbf{81.81} & 60.27 & \textbf{56.30} \\ 
    \bottomrule
    \end{tabularx}
    \label{tab:mlt17_task3}
    \vspace{-2mm}
\end{table}
Table \ref{tab:mlt17_task3} and Table \ref{tab:mlt19_task3} show the end-to-end language identification results on MLT17~\cite{nayef2017icdar2017} and MLT19~\cite{nayef2019icdar2019}, respectively. The proposed Multiplexed Mask TextSpotter achieves the best F-score (H-mean), precision, and average precision. Note that we didn't fine-tune our model on the MLT17 dataset for the MLT17 benchmark, which contains one fewer language (Hindi), but still managed to outperform existing methods in all metrics but recall.
Also, we implemented a post-processing step for the single-head Mask TextSpotter that infers the language from the recognized words similar to \cite{buvsta2018e2e,baek2020character}, and the results again show that the multiplexed model with an LPN outperforms the single-head model with post processing. This can be explained by the fact that our language prediction network infers the script based on visual cues directly, while the post-processing-based method could suffer from noisy text recognition results.

\begin{table}[t]
    \scriptsize
    \centering
    \caption{\textbf{Joint text detection and script identification results on MLT19.} All Task 3 numbers taken from the official ICDAR19-MLT website.}
    % \vspace{-2mm}
    \begin{tabularx}{0.9\linewidth}{lX*{3}X}
    \toprule
    Method & F & P & R & AP \\ 
    \midrule
    CRAFTS~\cite{baek2020character} & 68.34 & 78.52 & \textbf{60.50} & 53.75 \\
    \hline
    Single-head TextSpotter~\cite{liao2020mask} & 65.19 & 75.41 & 57.41 & 51.98 \\
    Multiplexed TextSpotter & \textbf{69.42} & \textbf{81.72} & 60.34 & \textbf{56.46} \\ 
    \bottomrule
    \end{tabularx}
    \label{tab:mlt19_task3}
   \vspace{-2mm}
\end{table}

\begin{table}[t]
    \scriptsize
    \centering
    \caption{\textbf{End-to-end recognition results on MLT19.} All numbers are from the official ICDAR19-MLT website except CRAFTS (paper), which comes from \cite{baek2020character}.}
    % \vspace{-2mm}
    \begin{tabularx}{0.9\linewidth}{lX*{2}X}
    \toprule
    Method & F & P & R \\ 
    \midrule
    E2E-MLT~\cite{buvsta2018e2e} & 26.5 & 37.4 & 20.5 \\
    RRPN+CLTDR~\cite{ma2018arbitrary} & 33.8 & 38.6 & 30.1 \\
    CRAFTS~\cite{baek2020character} & 51.7 & 65.7 & 42.7 \\
    CRAFTS (paper)~\cite{baek2020character} & \textbf{58.2} & \textbf{72.9} & \textbf{48.5} \\
    \hline
    Single-head TextSpotter~\cite{liao2020mask} & 39.7 & 71.8 & 27.4 \\ 
    Multiplexed TextSpotter & 48.2 & 68.0 & 37.3   \\ 
    \bottomrule
    \end{tabularx}
    \label{tab:mlt19_task4}
    \vspace{-2mm}
\end{table}

\subsection{End-to-end multilingual recognition task}

Table \ref{tab:mlt19_task4} shows the end-to-end multilingual recognition benchmark results on MLT19~\cite{nayef2019icdar2019}. Our method outperforms all methods except CRAFTS \cite{baek2020character}. We think that the difference in performance mainly comes from: (a) their ResNet-based feature extraction module with 24 Conv layers in their recognition head, as opposed to our feature extraction module with only 5 Conv layers, (b) the orientation estimation/link representation for vertical text that is common in East Asian languages, (c) the TPS-based rectification, and (d) the use ofthe ReCTS dataset \cite{zhang2019icdar} containing 20K additional training images. All these improvements are orthogonal to the proposed multiplexed framework and they can be combined. For example, the multiplexed model enables the potential improvement by allowing specific recognition heads to be customized to accommodate the vertical text. Regardless, we observe that the multiplexed model strongly outperforms the single-head model, demonstrating the effectiveness of the proposed multiplexed framework.
Figure \ref{fig:vis2_mlt19test} shows some qualitative visualization results of end-to-end recognition with the proposed Multiplexed TextSpotter on the MLT19 test set.
We see that the proposed model is able to successfully detect and recognize text from multiple languages within the same scene.

\section{Conclusion}

We propose a multiplexed network for end-to-end multilingual OCR. To our knowledge, this is the first end-to-end framework that trains text detection, segmentation, language identification and multiple text recognition heads in an end-to-end manner. The framework provides flexibility on freezing any part of the model and focus on training the other parts. The multiplexed pipeline is particularly useful when we need to support new languages, remove existing languages (e.g. for deployment in special scenarios that require only a subset of languages and/or have limited hardware resource), or improve/upgrade recognition model on certain languages, without worrying about harming the other existing languages of the model. We achieve state-of-the-art performance on the joint text detection and script identification tasks of both MLT19 and MLT17 benchmarks. Our experiments also show that with similar number of total parameters, the multiplexed model can achieve better results than a single unified recognition model with similar architectures. As future work, we plan to leverage task similarities~\cite{nguyen2020leep,tran2019transferability} to explore grouping related languages into single recognition heads. 

{\small
\bibliographystyle{ieee_fullname}
\bibliography{egbib}
}

\end{document}